\documentclass[runningheads]{llncs}
\usepackage[T1]{fontenc}
\usepackage{graphicx}
\usepackage{hyperref}
\usepackage{amsfonts}
\usepackage{multirow}
\usepackage{color}

\usepackage{colortbl}
\usepackage{framed,multirow}
\usepackage{xcolor}

\usepackage[misc]{ifsym} 
\begin{document}
\title{Transformer-based Annotation Bias-aware Medical Image Segmentation}

%
%
\author{Zehui Liao\inst{1} \and
Yutong Xie\inst{2} \and
Shishuai Hu\inst{1} \and
Yong Xia\inst{1,3,4}$^{(\textrm{\Letter})}$}
\authorrunning{Z. Liao et al.}
%
\institute{
National Engineering Laboratory for Integrated Aero-Space-Ground-Ocean Big Data Application Technology, School of Computer Science and Engineering, Northwestern Polytechnical University, Xi’an 710072, China \\
\email{yxia@nwpu.edu.cn}
\and
Australian Institute for Machine Learning, The University of Adelaide, Adelaide, SA, Australia.
\and
Ningbo Institute of Northwestern Polytechnical University, Ningbo 315048, China
\and
Research and Development Institute of Northwestern Polytechnical University in Shenzhen, Shenzhen 518057, China
}

\maketitle              

\begin{abstract}
Manual medical image segmentation is subjective and suffers from annotator-related bias, which can be mimicked or amplified by deep learning methods. Recently, researchers have suggested that such bias is the combination of the annotator preference and stochastic error, which are modeled by convolution blocks located after decoder and pixel-wise independent Gaussian distribution, respectively. It is unlikely that convolution blocks can effectively model the varying degrees of preference at the full resolution level. Additionally, the independent pixel-wise Gaussian distribution disregards pixel correlations, leading to a discontinuous boundary. This paper proposes a Transformer-based Annotation Bias-aware (TAB) medical image segmentation model, which tackles the annotator-related bias via modeling annotator preference and stochastic errors. TAB employs the Transformer with learnable queries to extract the different preference-focused features. This enables TAB to produce segmentation with various preferences simultaneously using a single segmentation head. Moreover, TAB takes the multivariant normal distribution assumption that models pixel correlations, and learns the annotation distribution to disentangle the stochastic error. We evaluated our TAB on an OD/OC segmentation benchmark annotated by six annotators. Our results suggest that TAB outperforms existing medical image segmentation models which take into account the annotator-related bias.

\end{abstract}

\section{Introduction}
Deep convolutional neural networks (DCNNs) have significantly advanced medical image segmentation~\cite{hesamian2019deep,tajbakhsh2020embracing,liu2021review}. 
However, their success relies heavily on accurately labeled training data~\cite{zhang2021understanding}, which are often unavailable for medical image segmentation tasks since manual annotation is highly subjective and requires the observer's perception, expertise, and concentration~\cite{wang2021annotation,fu2020retrospective,karimi2020deep,schaekermann2019understanding,joskowicz2019inter}. 
In a study of liver lesion segmentation using abdominal CT, three trained observers delineated the lesion twice over a one-week interval, resulting in the variation of delineated areas up to 10\% per observer and more than 20\% between observers~\cite{Suetens2017FundamentalsOM}.

To analyze the annotation process, it is assumed that a latent true segmentation, called meta segmentation for this study, exists as the consensus of annotators~\cite{Liao2021PADL,Zhang2020CMNet}. 
Annotators prefer to produce annotations that are different from the meta segmentation to facilitate their own diagnosis. 
Additionally, stochastic errors may arise during the annotation process. 
To predict accurate meta segmentation and annotator-specific segmentation, research efforts have been increasingly devoted to addressing the issue of annotator-related bias~\cite{ji2021learning,Zhang2020CMNet,Liao2021PADL}.

Existing methods can be categorized into three groups.
\textbf{\textit{Annotator decision fusion}}~\cite{ji2021learning,Mirikharaji2019noisyanno,Xiao2021Path} methods model annotators individually and use a weighted combination of multiple predictions as the meta segmentation.
Despite their advantages, these methods ignore the impact of stochastic errors on the modeling of annotator-specific segmentation~\cite{Jungo2018effect}.
\textbf{\textit{Annotator bias disentangling}}~\cite{Zhang2020experts,Zhang2020CMNet} methods estimate the meta segmentation and confusion matrices and generate annotator-specific segmentation by multiplying them.
Although confusion matrices characterize the annotator bias, its estimation is challenging due to the absence of ground truth.
Thus, the less-accurate confusion matrices seriously affect the prediction of meta segmentation.
\textbf{\textit{Preference-involved annotation distribution learning (PADL)}} framework ~\cite{Liao2021PADL} disentangles annotator-related bias into annotator preference and stochastic errors and, consequently, outperforms previous methods in predicting both meta segmentation and annotator-specific segmentation. 
Although PADL has recently been simplified by replacing its multi-branch architecture with dynamic convolutions~\cite{guo2022modeling}, it still has two limitations.
First, PADL uses a stack of convolutional layers after the decoder to model the annotator preference, which may not be effective in modeling the variable degrees of preference at the full resolution level, and the structure of this block, such as the number of layers and kernel size, needs to be adjusted by trial and error.
Second, PADL adopts the Gaussian assumption and learns the annotation distribution per pixel independently to disentangle stochastic errors, resulting in a discontinuous boundary.

To address these issues, we advocate extracting the features, on which different preferences focus.
Recently, Transformer~\cite{Vaswani2017trans,liu2021swin,liu2022swin} has drawn increasing attention due to its ability to model the long-range dependency. 
Among its variants, DETR~\cite{Carion2020DETR} has a remarkable ability to detect multiple targets at different locations in parallel, since it takes a set of different positional queries as conditions and focuses on features at different positions.
Inspired by DETR, we suggest utilizing such queries to represent annotators' preferences, enabling Transformer to extract different preference-focused features.
To further address the issue of missing pixel correlation, we suggest using a non-diagonal multivariate normal distribution~\cite{Monteiro2020SSN} to replace the pixel-wise independent Gaussian distribution.

In this paper, we propose a \textbf{T}ransformer-based \textbf{A}nnotation \textbf{B}ias-aware (\textbf{TAB}) medical image segmentation model, which can characterize the annotator preference and stochastic errors and deliver accurate meta segmentation and annotator-specific segmentation.
TAB consists of a CNN encoder, a Preference Feature Extraction (PFE) module, and a Stochastic Segmentation (SS) head.
The CNN encoder performs image feature extraction. 
The PFE module takes the image feature as input and produces $R+1$ preference-focused features in parallel for meta/annotator-specific segmentation under the conditions of $R+1$ different queries.
Each preference-focused feature is combined with the image feature and fed to the SS head.
The SS head produces a multivariate normal distribution that models the segmentation and annotator-related error as the mean and variance respectively, resulting in more accurate segmentation. 
We conducted comparative experiments on a public dataset (two tasks) with multiple annotators. Our results demonstrate the superiority of the proposed TAB model as well as the effectiveness of each component.

The main contributions are three-fold. 
(1) TAB employs Transformer to extract preference-focused features under the conditions of various queries, based on which the meta/annotator-specific segmentation maps are produced simultaneously.
(2) TAB uses the covariance matrix of a multivariate normal distribution, which considers the correlation among pixels, to characterize the stochastic errors, resulting in a more continuous boundary. 
(3) TAB outperforms existing methods in addressing the issue of annotator-related bias.

\section{Method}

\subsection{Problem Formalization and Method Overview}
Let a set of medical images annotated by $R$ annotators be denoted by $D=\left \{ x_{i}, y_{i1}, y_{i2}, \cdots, y_{iR} \right \}_{i=1}^{N}$, where $x_{i} \in \mathbb{R}^{C\times H\times W}$ represents the $i$-th image with $C$ channels and a size of $H\times W$, and $y_{ir} \in \left \{ 0,1 \right \} ^{K\times H \times W}$ is the annotation with $K$ classes given by the $r$-th annotator. 
We simplify the \textit{K}-class segmentation problem as \textit{K} binary segmentation problems.
Our goal is to train a segmentation model on $D$ so that the model can generate a meta segmentation map and $R$ annotator-specific segmentation maps for each input image.

\begin{figure}[t]
\centering
\includegraphics[width=\textwidth]{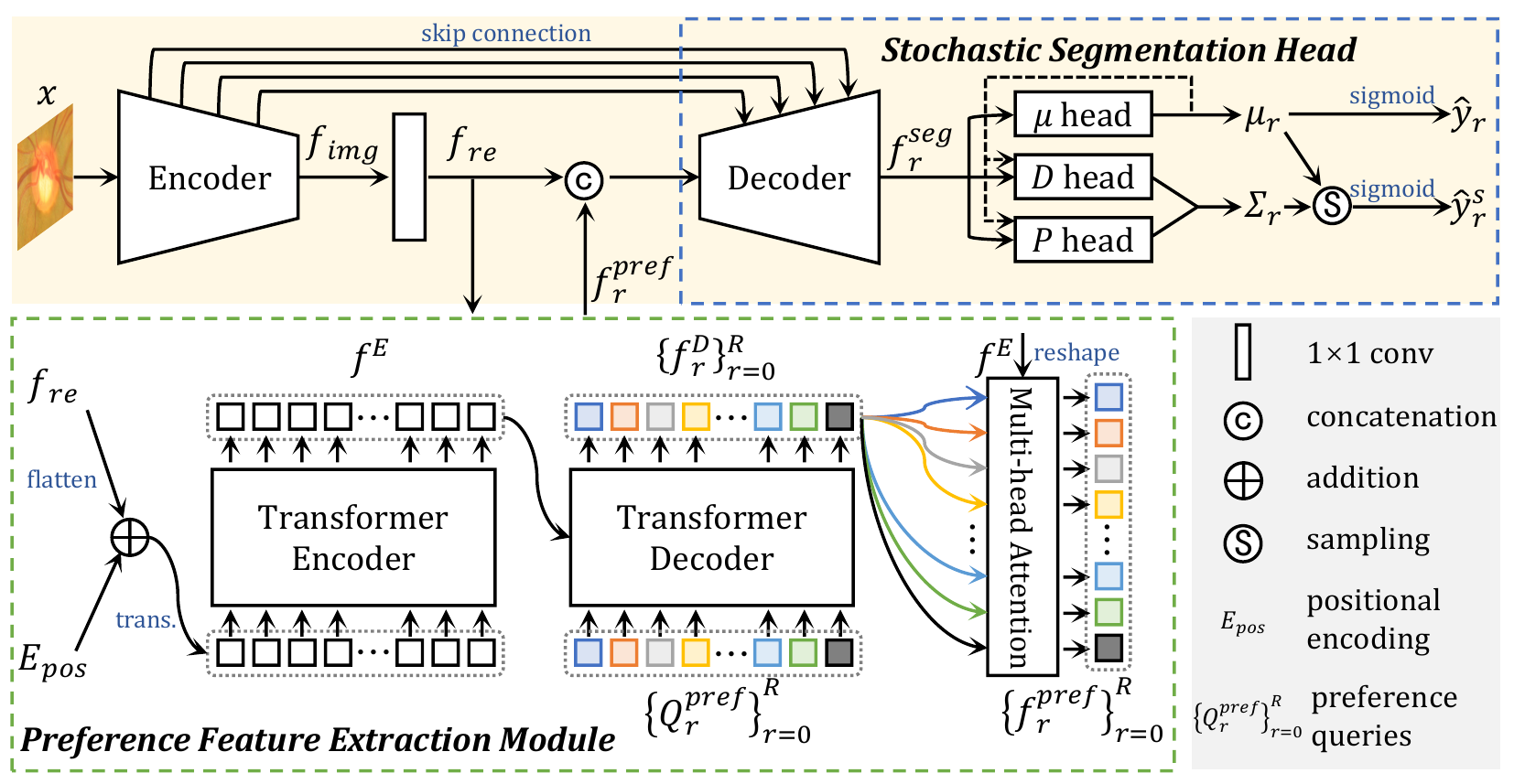}
\caption{Framework of TAB model, which consists of a CNN encoder, a PFE module, and a SS head.
`trans.' means the transpose operation.}
\label{fig: overview}
\end{figure}

Our TAB model contains three main components: a CNN encoder for image feature extraction, a PFE module for preference-focused feature production, and a SS head for segmentation prediction (see Fig.~\ref{fig: overview}). We now delve into the details of each component.

\subsection{CNN Encoder}
The ResNet34~\cite{he2016deep} pre-trained on ImageNet is employed as the CNN encoder. We remove its average pooling layer and fully connected layer to adjust it to our tasks. 
Skip connections are built from the first convolutional block and the first three residual blocks in the CNN encoder to the corresponding locations of the decoder in the SS head. 
The CNN encoder takes an image $x\in \mathbb{R}^{C \times H \times W}$ as its input and generates a high-level low-resolution feature map $f_{img}\in \mathbb{R}^{C'\times H'\times W'}$, where $C'=512$, $H'=\frac{H}{32}$, $W'=\frac{W}{32}$.
Moreover, $f_{img}$ is fed to a $1\times 1$ convolutional layer for channel reduction, resulting in $f_{re}\in \mathbb{R}^{d\times H'\times W'}$, where $d=256$.

\subsection{PFE Module}
The PFE module consists of an encoder-decoder Transformer and a multi-head attention block. 
Feeding the image feature $f_{re}$ to the PFE module, we have $R+1$ enhanced feature maps $\{ f_r^{pref} \}_{r=0}^{R}$, on which different preferences focus ($r=0$ for meta segmentation and others for $R$ annotator-specific segmentation). 
Note that meta segmentation is regarded as a special preference for simplicity.

\noindent \textbf{The Transformer Encoder} is used to enhance the image feature $f_{re}$.
The Transformer encoder consists of a multi-head self-attention module and a feed-forward network. 
Since the encoder expects a sequence as input, we collapse the spatial dimensions of $f_{re}$ and reshape it into the size of $d\times H'W'$.
Next, $f_{re}$ is added to the fixed positional encodings $E_{pos}$ and fed to the encoder.
The output of the Transformer encoder is denoted by $f^E$ and its size is $d\times H'W'$.

\noindent \textbf{The Transformer Decoder} accepts $f^E$ and $R+1$ learnable queries $\{ Q_r^{pref} \}_{r=0}^R$ of size $d=256$ as its input.
We aim to extract different preference-focused features based on the conditions provided by $\{ Q_r^{pref} \}_{r=0}^R$, which are called `preference queries' accordingly.
This decoder consists of a multi-head self-attention module for the intercommunication between queries, a multi-head attention module for feature extraction under the conditions of queries, and a feed-forward network.
And it produces $R+1$ features $\{ f_r^D \}_{r=0}^R$ of size $d=256$ in parallel.

\noindent \textbf{Multi-head Attention Block} has $m$ heads in it.
It takes the output of the Transformer decoder $\{ f_r^D \}_{r=0}^R$ as its input and computes multi-head attention scores of $f_r^D$ over the output of the encoder $f^E$, generating $m$ attention heatmaps per segmentation.
The output of this block is denoted as $\{ f_r^{pref} \}_{r=0}^R$, and the size of $f_r^{pref}$ is $m \times H' \times M'$.
Then, $\{ f_r^{pref} \}_{r=0}^R$ are individually decoded by SS head, resulting in $R+1$ different preference-involved segmentation maps.

\subsection{SS Head}
The SS head aims to disentangle the annotator-related bias and produce meta and annotator-specific segmentation maps.
Given an input image, we assume that the annotation distribution over annotators follows a multivariant normal distribution.
Thus, we can learn the annotation distribution and disentangle the annotator-related bias by modeling it as the covariance matrix that considers pixel correlation.
First, we feed the concatenation of $f_{re}$ and $f_r^{pref}$ to a CNN decoder, which is followed by batch normalization and ReLU, and obtain the feature map $f_{r}^{seg} \in \mathbb{R}^{32\times H\times W}$.
Second, we establish the multivariant normal distribution $\mathcal{N}(\mu (x), \Sigma (x))$ via predicting the $\mu (x) \in \mathbb{R}^{K \times H \times W}$ and $\Sigma (x) \in \mathbb{R}^{{(K \times H\times W)}^2}$ based on $f_r^{seg}$.
However, the size of the covariance matrix scales with ${(K \times H\times W)}^2$, making it computationally intractable. To reduce the complexity, we adopt the low-rank parameterization of the covariance matrix~\cite{Monteiro2020SSN}
\begin{equation}
    \Sigma = P\times P^T + D
\label{eq: Sigma}
\end{equation}
where $P \in \mathbb{R}^{(K \times H\times W) \times \alpha}$ is the covariance factor, $\alpha$ defines the rank of the parameterization, $D$ is a diagonal matrix with $K \times H\times W$ diagonal elements. 
We employ three convolutional layers with $1\times 1$ kernel size to generate $\mu (x)$, $P$, and $D$, respectively.
In addition, the concatenation of $\mu (x)$ and $f_r^{seg}$ is fed to the $P$ head and $D$ head, respectively, to facilitate the learning of $P$ and $D$.
Finally, we get $R+1$ distributions. Among them,
$\mathcal{N}(\mu_{MT} (x), \Sigma_{MT} (x))$ is for meta segmentation and $\mathcal{N}(\mu_{r} (x), \Sigma_{r} (x)), r=1,2,...,R$ are for annotator-specific segmentation.
The probabilistic meta/annotator-specific segmentation map ($\hat{y}_{MT}$/$\hat{y}_r$) is calculated by applying the sigmoid function to the estimated $\mu_{MT}$/$\mu_{r}$.
We can also produce the probabilistic annotator bias-involved segmentation maps $\hat{y}_{MT}^s$/$\hat{y}_r^s$ by applying the sigmoid function to the segmentation maps sampled from the established distribution $\mathcal{N}(\mu_{MT} (x), \Sigma_{MT} (x))$/$\mathcal{N}(\mu_{r} (x), \Sigma_{r} (x))$.

\subsection{Loss and Inference}
The loss of our TAB model contains two items: the meta segmentation loss $L_{MT}$ and annotator-specific segmentation loss $L_{AS}$, shown as follows
\begin{equation}
    L = L_{MT}(y^s, \hat{y}_{MT}^s) + \sum_{r=1}^R{L}_{AS}(y_r, \hat{y}_r^s)
\end{equation}
where $L_{MT}$ and $L_{AS}$ are the binary cross-entropy loss, $y^s$ is a randomly selected annotation per image,  $y_r$ is the delineation given by annotator $A_r$. 

During inference, the estimated probabilistic meta segmentation map $\hat{y}_{MT}$ is evaluated against the mean voting annotation.
The estimated probabilistic annotator-specific segmentation map $\hat{y}_r$ is evaluated against the annotation $y_r$ given by the annotator $A_r$.

\section{Experiments and Results}
\subsection{Dataset and Experimental Setup}

\noindent\textbf{Dataset.} The RIGA dataset~\cite{almazroa2017agreement} is a public benchmark for optic disc and optic cup segmentation, which contains 750 color fundus images from three sources, including 460 images from MESSIDOR, 195 images from BinRushed, and 95 images from Magrabia. 
Six ophthalmologists from different centers labeled the optic disc/cup contours manually on each image. 
We use 655 samples from BinRushed and MESSIDOR for training and 95 samples from Magrabia for test~\cite{yu2019robust,ji2021learning,Liao2021PADL}.
The 20\% of the training set that is randomly selected is used for validation.

\noindent \textbf{Implementation Details.} All images were normalized via subtracting the mean and dividing by the standard deviation. The mean and standard deviation were counted on training cases.
We set the batch size to 8 and resized the input image to $256\times256$.
The Adam optimizer~\cite{Kingma2015AdamAM} with default settings was adopted.
The learning rate $lr$ was set to $5e-5$ and decayed according to the polynomial policy $lr = lr_{0} \times \left ( 1-t/T  \right ) ^{0.9}$, where $t$ is the epoch index and $T=300$ is the maximum epoch. 
All results were reported over three random runs. Both mean and standard deviation are given.

\noindent \textbf{Evaluation Metrics.} We adopted the Soft Dice ($D^s$) as the performance metric. 
At each threshold level, the Hard Dice is calculated between the segmentation and annotation maps.
Soft Dice is calculated via averaging the hard Dice values obtained at multiple threshold levels, \textit{i.e.}, (0.1, 0.3, 0.5, 0.7, 0.9) for this study. 
Based on the Soft Dice, there are two performance metrics, namely \textit{Average} and \textit{Mean Voting}. 
\textit{Mean Voting} is the Soft Dice between the predicted meta segmentation and the mean voting annotation. 
A higher \textit{Mean Voting} represents better performance on modeling the meta segmentation. 
The annotator-specific predictions are evaluated against each annotator’s delineations, and the average Soft Dice of all annotators is denoted as \textit{Average}. 
A higher \textit{Average} represents better performance on mimicking all annotators.

\begin{table*}[h]
\footnotesize
\renewcommand\arraystretch{0.9}
\setlength\tabcolsep{0.8pt}
\centering
\caption{
Performance ($\mathcal{D}_{disc}^{s}$ (\%), $\mathcal{D}_{cup}^{s}$ (\%)) of our TAB, seven competing models, and two variants of TAB on the RIGA dataset. 
From left to right: Performance in mimicking the delineations of each annotator ($A_r$, \textit{r}=1, 2, ..., 6), \textit{Average}, and \textit{Mean Voting}.
The standard deviation is shown as the subscript of the mean.
Except for two variants of TAB and the `Multi-Net' setting ($M_r$), the best results in \textit{Average} and \textit{Mean Voting} columns are highlighted in \textcolor{blue}{blue}.
}
\begin{tabular}{l|c|c|c|c}
\hline \hline
Methods      & $A_1$            & $A_2$            & $A_3$           & $A_4$             \\ \hline
$M_r$        & $96.20_{0.05}$,$84.43_{0.16}$ & $95.51_{0.02}$,$84.81_{0.36}$ & $96.56_{0.04}$,$83.15_{0.11}$ & $96.80_{0.03}$,$87.89_{0.09}$ \\ \hline
MH-UNet      & $96.25_{0.29}$,$83.31_{0.49}$ & $95.27_{0.10}$,$81.82_{0.24}$ & $96.72_{0.28}$,$77.32_{0.35}$ & $97.00_{0.11}$,$88.03_{0.29}$  \\ 
MV-UNet      & $95.11_{0.02}$,$76.85_{0.34}$ & $94.53_{0.05}$,$78.45_{0.44}$ & $95.57_{0.03}$,$77.68_{0.22}$ & $95.70_{0.02}$,$76.27_{0.56}$  \\ \hline
MR-Net       & $95.33_{0.46}$,$81.96_{0.47}$ & $94.72_{0.43}$,$81.13_{0.78}$ & $95.65_{0.14}$,$79.04_{0.81}$ & $95.94_{0.03}$,$84.13_{2.61}$  \\ 
CM-Net       & $96.26_{0.07}$,$84.50_{0.14}$ & $95.41_{0.08}$,$81.46_{0.52}$ & $96.55_{0.88}$,$81.80_{0.41}$ & $96.80_{1.17}$,$87.50_{0.51}$  \\ 
PADL         & $96.43_{0.03}$,$85.21_{0.26}$ & $95.60_{0.05}$,$85.13_{0.25}$ & $96.67_{0.02}$,$82.74_{0.37}$ & $96.88_{0.11}$,$88.80_{0.10}$  \\ 
AVAP         & $96.20_{0.13}$,$85.79_{0.59}$ & $95.44_{0.03}$,$84.50_{0.51}$ & $96.47_{0.03}$,$81.65_{0.95}$ & $96.82_{0.02}$,$89.61_{0.20}$  \\ \hline
Ours         & $96.32_{0.05}$,$86.13_{0.18}$ & $96.21_{0.07}$,$85.90_{0.17}$ & $96.90_{0.08}$,$84.64_{0.14}$ & $97.01_{0.07}$,$89.40_{0.12}$  \\ \hline
Ours{\fontsize{1pt}{\baselineskip}\selectfont{w/o PFE}} & $95.67_{0.03}$,$81.69_{0.13}$ & $95.12_{0.02}$,$80.16_{0.11}$ & $96.08_{0.04}$,$79.42_{0.16}$ & $96.40_{0.02}$,$78.53_{0.17}$  \\
Ours{\fontsize{1pt}{\baselineskip}\selectfont{w/o SS}}  & $96.39_{0.11}$,$84.82_{0.46}$ & $95.55_{0.02}$,$83.68_{0.47}$ & $96.41_{0.05}$,$82.73_{0.29}$ & $96.77_{0.07}$,$88.21_{0.47}$  \\\hline \hline

Methods      &$A_5$             & $A_6$                  & \textit{Average}    & \textit{Mean Voting} \\ \hline
$M_r$        & $96.70_{0.03}$,$83.27_{0.15}$ & $97.00_{0.00}$,$80.45_{0.01}$ & $96.46_{0.03}$,$84.00_{0.16}$ & / \\ \hline
MH-UNet      & $97.09_{0.17}$,$78.69_{0.90}$ & $96.82_{0.41}$,$75.89_{0.32}$ & $96.54_{0.09}$,$80.84_{0.41}$ & $97.39_{0.06}$,$85.27_{0.09}$ \\  
MV-UNet      & $95.85_{0.04}$,$78.64_{0.20}$ & $95.62_{0.02}$,$74.74_{0.15}$ & $95.40_{0.02}$,$77.11_{0.28}$ & $97.45_{0.04}$,$86.08_{0.12}$ \\ \hline 
MR-Net       & $95.87_{0.20}$,$79.00_{0.25}$ & $95.71_{0.06}$,$76.18_{0.27}$ & $95.54_{0.19}$,$80.24_{0.70}$ & $97.50_{0.07}$,$87.21_{0.19}$ \\  
CM-Net       & $96.28_{1.06}$,$82.43_{0.25}$ & $96.85_{1.39}$,$78.77_{0.33}$ & $96.36_{0.06}$,$82.74_{0.55}$ & $96.41_{0.34}$,$81.21_{0.12}$ \\  
PADL         & $96.80_{0.08}$,$83.42_{0.39}$ & $96.89_{0.02}$,$79.76_{0.36}$ & $96.53_{0.01}$,$84.32_{0.06}$ & $97.77_{0.03}$,$87.77_{0.08}$ \\ 
AVAP         & $96.29_{0.04}$,$83.63_{0.35}$ & $96.68_{0.02}$,$80.06_{0.48}$ & $96.32_{0.13}$,$84.21_{0.46}$ & $\textcolor{blue}{97.86}_{0.02}$,$87.60_{0.27}$ \\ \hline
Ours         & $96.80_{0.06}$,$84.50_{0.13}$ & $96.99_{0.04}$,$82.55_{0.16}$ & $\textcolor{blue}{96.70}_{0.04}$,$\textcolor{blue}{85.52}_{0.06}$ & $97.82_{0.04}$,$\textcolor{blue}{88.22}_{0.07}$  \\ \hline
Ours{\fontsize{1pt}{\baselineskip}\selectfont{w/o PFE}} & $96.40_{0.04}$,$79.84_{0.15}$ & $96.25_{0.01}$,$75.69_{0.16}$ & $95.99_{0.05}$,$79.22_{0.14}$ & $97.71_{0.04}$,$87.51_{0.14}$  \\
Ours{\fontsize{1pt}{\baselineskip}\selectfont{w/o SS}}  & $96.63_{0.09}$,$82.88_{0.60}$ & $96.81_{0.06}$,$79.90_{0.52}$ & $96.43_{0.03}$,$83.71_{0.37}$ & $97.18_{0.02}$,$85.36_{0.67}$  \\\hline \hline
\end{tabular}
\label{tab:RIGA-comparision}
\end{table*}

\subsection{Comparative Experiments}
\begin{figure}[t]
\centering
\includegraphics[width=\textwidth]{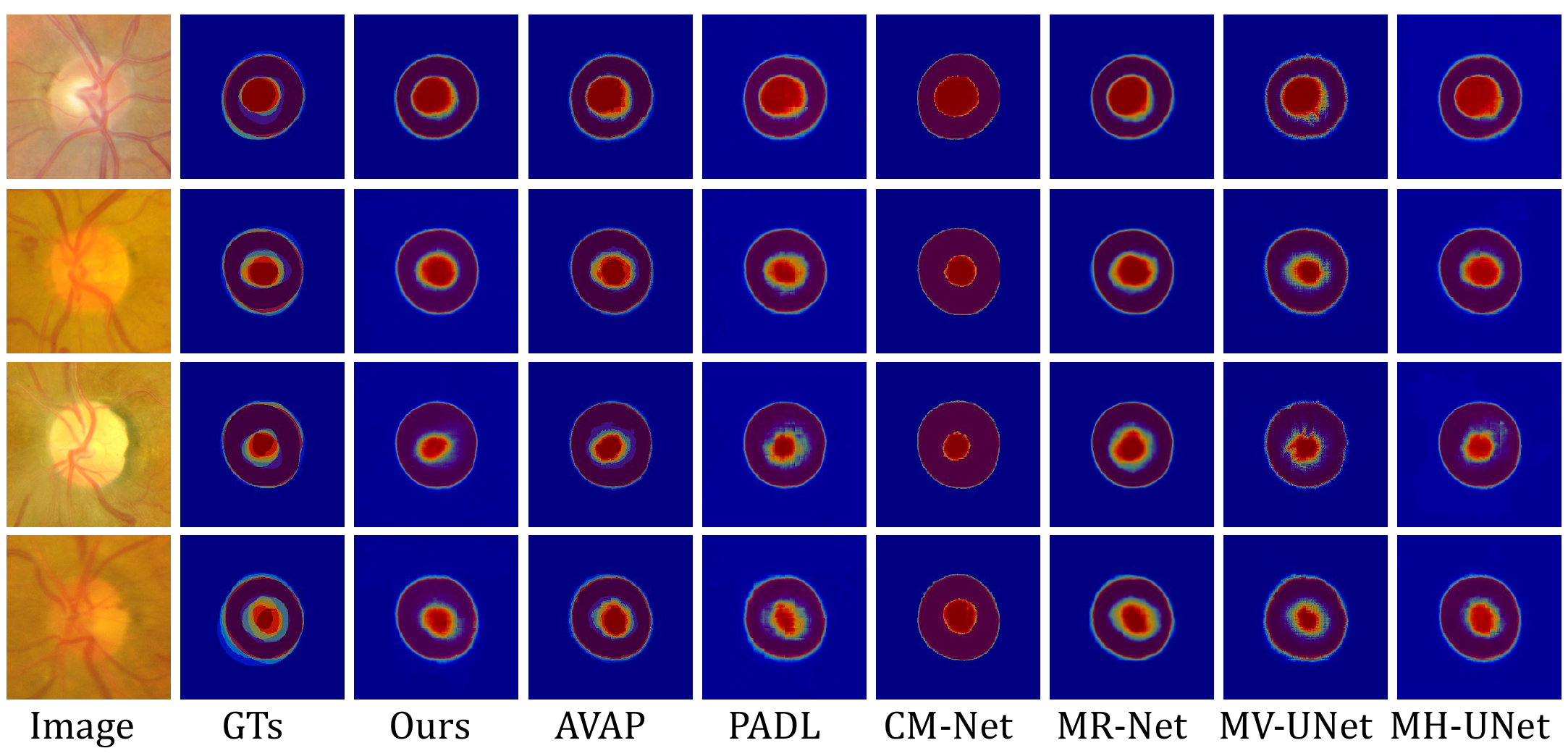}
\caption{Visualization of predicted meta segmentation maps obtained by applying six competing methods and our TAB to four cases from the RIGA dataset, together with ground truths (GTs, $i.e.$, mean voting annotations).
}
\label{fig: visualization}
\end{figure}

We compared our TAB to three baseline models and four recent segmentation models that consider the annotator-bias issue, including 
(1) the baseline `Multi-Net' setting, under which $R$ U-Nets (denoted by $M_r, r=1,2,...,R$, $R=6$ for RIGA) were trained and tested with the annotations provided by annotator $A_r$, respectively;
(2) MH-UNet~\cite{Guan2018whosaidwhat}: a U-Net variant with multiple heads, each accounting for imitating the annotations from a specific annotator; 
(3) MV-UNet~\cite{Falk2019unet}: a U-Net trained with the mean voting annotations;
(4) MR-Net~\cite{ji2021learning}: an annotator decision fusion method that uses an attention module to characterize the multi-rater agreement;
(5) CM-Net~\cite{Zhang2020CMNet}: an annotator bias disentangling method that uses a confusion matrix to model human errors;
(6) PADL~\cite{Liao2021PADL}: a multi-branch framework that models annotator preference and stochastic errors simultaneously;
and (7) AVAP~\cite{guo2022modeling}: a method that uses dynamic convolutional layers to simplify the multi-branch architecture of PADL.
The soft Dice of optic disc $\mathcal{D}_{disc}^s$ and optic cup $\mathcal{D}_{cup}^s$ obtained by our model and competing methods were listed in Table~\ref{tab:RIGA-comparision}.
It shows that our TAB achieves the second highest $\mathcal{D}_{disc}^s$ and highest $\mathcal{D}_{cup}^s$ on \textit{Mean Voting}, and achieves the highest $\mathcal{D}_{disc}^s$ and $\mathcal{D}_{cup}^s$ on \textit{Average}.
We also visualize the segmentation maps predicted by TAB and other competing methods (see Fig.~\ref{fig: visualization}). It reveals that TAB can produce the most accurate segmentation map compared to the ground truth.

\subsection{Ablation Analysis}
Both the PFE module and SS head play an essential role in the TAB model, characterizing the annotators' preferences and stochastic errors independently.
To evaluate the contributions of them, we compared TAB with two of its variants, $i.e.$, `Ours w/o PFE’ and `Ours w/o SS’.
In `Ours w/o SS’, the SS head is replaced by the CNN decoder, which directly converts image features to a segmentation map.
`Ours w/o PFE' contains a CNN encoder and an SS head. This variant is, of course, not able to model the preference of each annotator.
The performance of our TAB and its two variants was given in Table~\ref{tab:RIGA-comparision}.
It reveals that, when the PFE module was removed, the performance in reconstructing each annotator’s delineations drops obviously. 
Meanwhile, without the SS head, \textit{Mean Voting} score drops from 97.82\% to 97.18\% for optic disc segmentation and from 88.22\% to 85.36\% for optic cup segmentation.
It indicates that the PFE module contributes substantially to the modeling of each annotator’s preference, and the SS head can effectively diminish the impact of stochastic errors and produce accurate meta/annotator-specific segmentation.

\begin{table}[t]
\caption{Performance of the TAB with complete SS head and its three variants on the RIGA dataset. The \textit{Average} and \textit{Mean Voting} ($\mathcal{D}_{disc}^s$(\%),$\mathcal{D}_{cup}^s$(\%)) are used as the performance metrics. The standard deviation is shown as the subscript of the mean.}
\label{tab: SSH_RIGA}
\centering
\footnotesize
\renewcommand\arraystretch{0.9}
\begin{tabular}{c|c|c|c|c}
\hline \hline
 $\mathcal{N}(\mu,\sigma)$ & $\mathcal{N}(\mu,\Sigma)$ & $\mu$ prior & \textit{Average}                              & \textit{Mean Voting }\\ \hline
          &            &                                             & $96.43_{0.03}$, $83.71_{0.37}$       & $97.18_{0.02}$, $85.36_{0.67}$ \\ 
 $\surd$  &            &                                             & $96.54_{0.02}$, $84.56_{0.09}$       & $97.66_{0.02}$, $87.18_{0.12}$ \\ 
          & $\surd$    &                                             & $96.67_{0.06}$, $85.24_{0.12}$       & $97.76_{0.07}$, $87.87_{0.09}$ \\ 
          & $\surd$    & $\surd$                                     & $96.70_{0.04}$, $85.52_{0.06}$       & $97.82_{0.04}$, $88.22_{0.07}$ \\ \hline \hline
\end{tabular}
\end{table}

\noindent \textbf{Analysis of the SS Head.}
The effect of each component in the SS head was accessed using \textit{Average} and \textit{Mean Voting}. Table~\ref{tab: SSH_RIGA} gives the performance of the TAB with complete SS head and its three variants.
We compare the stochastic errors modeling capacity of multivariate normal distribution $\mathcal{N}(\mu, \Sigma)$ and the pixel-wise independent Gaussian distribution $\mathcal{N}(\mu, \sigma)$~\cite{Liao2021PADL}. 
Though both $\mathcal{N}(\mu, \Sigma)$ and $\mathcal{N}(\mu, \sigma)$ can reduce the impact of stochastic errors, $\mathcal{N}(\mu, \Sigma)$ performs better than $\mathcal{N}(\mu, \sigma)$ on the test set.
We also explored the influence of $\mu$ prior.
It reveals that the $\mu$ prior can facilitate the learning of $\mathcal{N}(\mu, \Sigma)$ and improve the capacity of meta/annotator-specific segmentation.

\section{Conclusion}
In this paper, we propose the TAB model to address the issue of annotator-related bias that existed in medical image segmentation.
TAB leverages the Transformer with multiple learnable queries on extracting preference-focused features in parallel and the multivariate normal distribution on modeling stochastic annotation errors.
Extensive experimental results on the public RIGA dataset with annotator-related bias demonstrate that TAB achieves better performance than all competing methods.

\vspace{10px}
\noindent \textbf{Acknowledgment.} This work was supported in part by the National Natural Science Foundation of China under Grant 62171377, in part by the Key Technologies Research and Development Program under Grant 2022YFC2009903 / 2022YFC2009900, in part by the Natural Science Foundation of Ningbo City, China, under Grant 2021J052, and in part by the Innovation Foundation for Doctor Dissertation of Northwestern Polytechnical University under Grant CX2022056.

\bibliographystyle{splncs04}
\bibliography{mybib}

\end{document}